# A Multi-Head Convolutional Neural Network With Multi-path Attention improves Image Denoising

Jiahong Zhang[1,2] , Meijun Qu[3], Ye Wang[1,2], and Lihong Cao[1,2 (B)]

[1] State Key Laboratory of Media Convergence and Communication, Communication University of China, Beijing, China.
[2] Neuroscience and Intelligent Media Institute, Communication University of China, Beijing, China.
zhangjh@cuc.edu.cn, yewang@cuc.edu.cn, lihong.cao@cuc.edu.cn
[3] School of Information and Communication Engineering, Communication University of China, Beijing, China.
qumeijun@cuc.edu.cn

**Abstract.** Recently, convolutional neural networks (CNNs) and attention mechanisms have been widely used in image denoising and achieved satisfactory performance. However, the previous works mostly use a single head to receive the noisy image, limiting the richness of extracted features. Therefore, a novel CNN with multiple heads (MH) named MHCNN is proposed in this paper, whose heads will receive the input images rotated by different rotation angles. MH makes MHCNN simultaneously utilize features of rotated images to remove noise. To integrate these features effectively, we present a novel multi-path attention mechanism (MPA). Unlike previous attention mechanisms that handle pixel-level, channel-level, or patch-level features, MPA focuses on features at the image level. Experiments show MHCNN surpasses other state-of-the-art CNN models on additive white Gaussian noise (AWGN) denoising and real-world image denoising. Its peak signal-to-noise ratio (PSNR) results are higher than other networks, such as BRDNet, RIDNet, PAN-Net, and CSANN. The code is accessible at https://github.com/JiaHongZ/MHCNN.

## 1 Introduction

Due to various problems in image acquisition equipment, the collected images often contain noise that can not be ignored. Image denoising aims to generate a clean image $x$ from a given noisy image $y$, modeled as $y = x + v$. Here $v$ denotes the noise, and AWGN is commonly used. Recently, CNNs achieved remarkable results in this task. Compared to traditional denoising methods [5], CNNs can be trained end-to-end, which are easy to optimize and have better denoising results.

Zhang et al. [35] utilized residual learning and batch normalization to construct DnCNN for AWGN denoising, super-resolution, and JPEG deblocking. Then, [34] were proposed to deepen the network by residual connections and got better results than DnCNN. Some works such as BRDNet [29] and DHDN [24] showed that widening the network can also improve the denoising performance. However, these methods used more convolution layers, increasing the computational complexity. Therefore, it is essential to extract features effectively.

The attention mechanism is commonly used to increase the CNN's capacity and flexibility of extracting image features. This paper classifies the previous attention mechanisms into three types: pixel-level, channel-level, and patch-level. For pixel-level attention, the non-local operation is a classical method in image restoration. It makes full use of the information from adjacent pixels and achieves success in image restoration [20], image resolution [7] and image denoising [33, 23]. Channel-level attention weights each feature channel and pays more attention to those important channels so that it improves denoising performance [2]. CSANN [31] and MRSNet [19] combined pixel-level attention with channel-level attention to get better denoising results than that of single attention. For patch-level attention, it can establish the relation between image patches and achieve good results in image enhancement [3].

Although these methods mentioned above achieved high performance, only one input head limits their abilities of extracting full features. This paper suggests that considering multiple rotation angles of the input image will get better results than one angle. It is different from the rotation pre-processing of data enhancement. Data enhancement can make CNNs learn the translated image features, but these features can not be simultaneously used because of the single input head limitation. We take a step to propose a multi-head convolutional neural network (MHCNN) with multi-path attention (MPA), which achieves state-of-the-art results. The different heads of MHCNN receive the input images rotated around the center to obtain rich features. MPA will integrate these features from different CNN heads to remove noise effectively, which is quite different from the previous attention mechanisms because MPA is image-level. The superiority of the proposed MHCNN is described in Section 4 and 5.

The main contributions of this paper are as follows:

(1) We propose a novel denoising network MHCNN. Its multiple heads (MH) will utilize features from multiple rotation angles of the input image.

(2) A novel attention module MPA is proposed to integrate the features from the different CNN heads. MPA focuses on image-level features rather than previous pixel-level, channel-level, and patch-level features.

(3) Ablation experiments show that the proposed MH with MPA mechanism is pluggable and can improve the denoising performance of the single head model.

(4) The proposed MHCNN can achieve state-of-the-art AWGN and real-world image denoising.

## 2   Related work

Image denoising has received extensive attention for its indispensable role in many practical applications. This paper focuses on the CNN-based image denoising.

### 2.1   CNNs for image denoising

In [16], Jain et al. claimed that CNNs have similar or even better representation power than traditional denoising models. Then, Zhang et al. [35] achieved fast and stable training and good denoising performance by integrating the residual learning and batch normalization to CNN. Singh et al. [27] used ResNet blocks to construct the network and get better results than common convolution layers. In addition to deepening the network, widening the network is also an effective way. BRDNet [29] is two-path networks and get better results than the single-path networks. U-Net-based networks such as MWCNN [21] and DHDN [24] are three-path network architecture. They further improved the denoising performance. However, these methods only consider one angle of the input image, resulting in extracting insufficient features. The proposed MHCNN uses MH to receive image features of multiple rotation angles to solve this problem.

### 2.2   Attention mechanisms for image denoising

Only MH is not enough because features from these heads need effective processing. Using attention mechanism is a popular method to increase this ability of CNNs. For pixel-level, non-local attention is commonly used, which restores the damaged pixel using its neighbors [20,33]. Some other pixel-level attention mechanisms use attention to guide the previous stage for image denoising. In ADNet [28], this guidance is achieved by convolution and multiplation operation. In PAN-Net [23], proportional-integral-derivative (PID) is used to get the guidance. Channel-level attention in RIDNet [2] utilized the relationship between the channel features to exploit and learn the critical content of the image. It achieves satisfying results both on AWGN and real-world image denoising. Patch-level attention is often used in vision transformer, which establishes the connection between image patches [8,3]. These attention methods also have the limitation of a single head. We propose a novel image-level MPA, which effectively integrate the features from the different heads of MHCNN.

## 3   The Proposed Method

### 3.1   Network architecture

Fig. 1 shows the proposed MHCNN. Given a noisy image, we firstly rotate it by 0°, 90°, and 180°to construct the inputs of the three heads. The main body of MHCNN consists of three parts as follows:

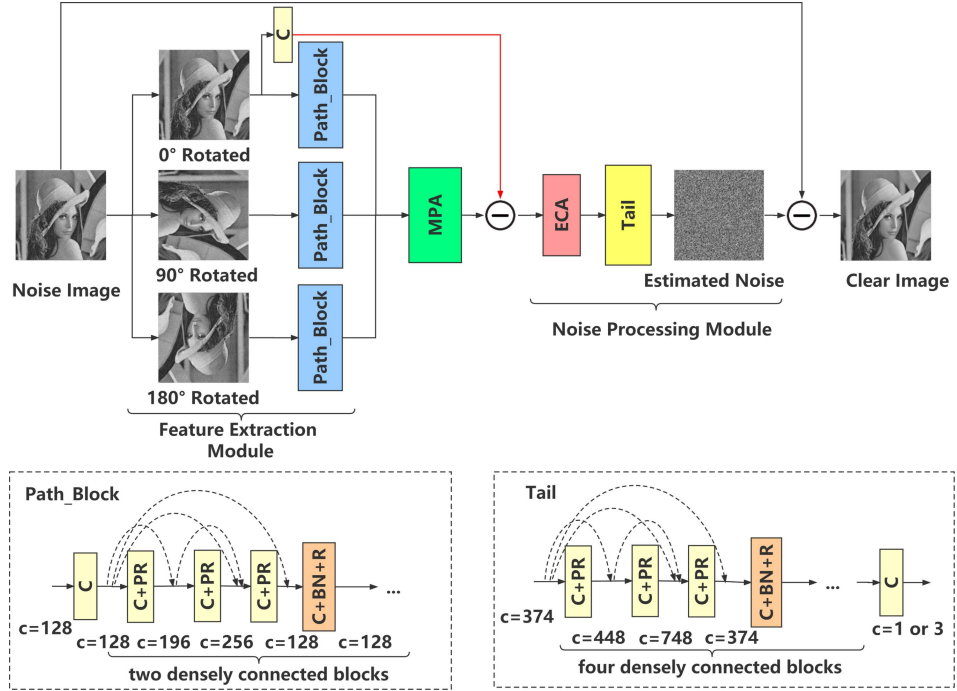

**Fig. 1.** The architecture of MHCNN. The feature extraction module consists of three heads to receive the input images with different rotation angles. Path_block is used to extract features from the different heads. MPA will integrate the features and get noise maps by the residual connection. At the end of MHCNN, noise processing module makes a further process to noise maps and genarates the estimated noise, where ECA is the effective channel attention layer [30]. Path_block and Tail are made up of densely connected blocks.

**Multi-head feature extraction module** Traditional CNN denoisers have a single head to receive the noisy image, which extracts limited features. We introduce MH that utilizes the features from noisy images with different rotation angles. For each head, we use Path_block shown in Fig. 1 to extract features. Path_block is a variant of Densnet block [13]. We first execute a 1×1 convolution operation to generate 128 feature maps. These feature maps are processed by two densely connected blocks composed of three C+PR followed by one C+BN+R. Here C is convolution layer, BN is batch normalization [15], R is ReLU [18] and PR is parametric rectified linear unit [11]. Convolution layers in densely connected blocks are set kernel size 3×3, stride 1, and padding 1. Ablation experiments in Section 5 show that MH improves the denoising performance.

**MPA** MH extracts features from multiple rotation angles of the input image. We assume that the rotated images contain many common features with the original one. In contrast, whether the image is rotated or not, the noise is independent. MPA module is designed to integrate the common features from different CNN heads, shown in Fig. 2. The input of MPA consists of *x*, *xr*1, *xr*2, which corresponds to the features of 0°, 90°, 180°rotated images, respectively. These three inputs have the same size (*b, c, h, w*), where b is the batch size, c is the number of channels, h and w are the height and width of the features. We use 1×1 convolution to process the input so that it endows MPA the learning ability. After the convolution operations, there are three data flows, *xr*1, *x*, and *xr*2. For *xr*1 flow, we reshape *x* to (*b, h, c, w*) and *xr*1 to (*b, h, w, c*), and then multiply them by matrix multiplication. This multiplication projects *x* to space of *xr*1, which gets a fusion representation of the image and rotated image. Then, it is normalized by instance norm (IN) [14] and is reshaped to (*b, h, c, c*). We subsequently project the (*b, h, c, c*) fusion representation to the space of *x* by multiplying it with another (*b, h, c, w*) *x*. Normalization and resizing are used again, and we finally project the *xr*1 to *x*. According to Fig. 2, the projection *xr*1_*p* can be represented as follows:

$$xr1\_p = Reshape(IN(Reshape(IN(Reshape(Conv(x)) \\ \times Reshape(Conv(xr1)))) \times Reshape(Conv(x))))) \quad (1)$$

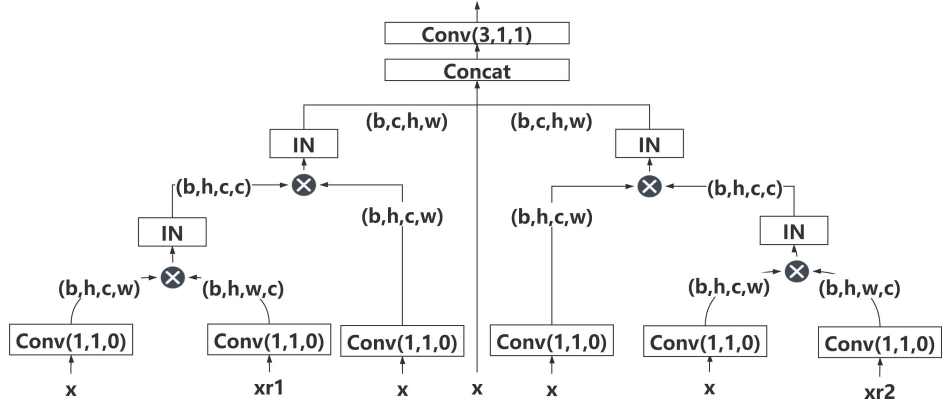

**Fig. 2.** The architecture of MPA module.

The projection from *xr*2 to *x* denoted as *xr*2_*p* is similar to *xr*1_*p*. For flow *x*, it remains the same. At the end of MPA, we concatenate these three data flows together. Thus the output of MPA is as follows:

$$MPA(xr1, xr2, x) = Conv(Concat(xr1\_p, xr2\_p, x)) \quad (2)$$

MPA projects the rotated images onto the original image through transformation. This image-level operation can effectively extract image features. The relevant analysis is in Section 5.2.

**Noise processing module** The output of MPA is the integrated features containing 128×3 channels. We subtract these features from the 0°rotated image to obtain noise maps, which are used as the input of the noise processing module.

**Table 1.** PSNR (dB) results for different networks on Set12 with noise levels of 15, 25, 50.

| Methods | $\sigma = 15$ | $\sigma = 25$ | $\sigma = 50$ |
|---|---|---|---|
| BM3D[5] | 32.37 | 29.97 | 26.72 |
| DnCNN[35] | 32.86 | 30.43 | 27.18 |
| FFDNet[36] | 32.77 | 30.44 | 27.32 |
| ADNet[28] | 32.98 | 30.58 | 27.37 |
| BRDNet[29] | 33.03 | 30.61 | 27.45 |
| CSANN[31] | – | 30.72 | 27.64 |
| PAN-Net[23] | 33.14 | **30.90** | 27.58 |
| MHCNN | **33.21** | 30.84 | **27.70** |

Here we use a 1×1 convolution layer to get 128×3 features from the 0°rotated image. The effective channel attention (ECA) layer [30] is used to weigh the different channels, and the followed Tail block will further process the noise maps, which is shown in Fig. 1. Tail block consists of four densely connected blocks, and its output is the estimated noise with 1 channel for gray images or 3 channels for color images. After Tail, we subtract the estimated noise from the noisy image to generate the clear image.

Given an input, MHCNN can be represented as:

$$\begin{aligned} x &= Path\_block(input) \\ xr1 &= Path\_block(Rot90°(input)) \\ xr2 &= Path\_block(Rot180°(input)) \\ noise &= input - MPA(xr1, xr2, x) \\ output &= input - Tail(ECA(noise)) \end{aligned} \quad (3)$$

We use l2 loss as the loss function of the proposed MHCNN, where $y_i$ is the real clear image, $output_i$ is the predicted clear image, and $N$ is the number of training samples:

$$L(\Theta) = \frac{1}{2N} \sum_n (y_i - output_i)^2, i \in [1, N] \quad (4)$$

### 3.2 Training setting

This model is implemented by python 3.5, PyTorch 1.5.1 with Cuda 9.2. The Adam[17] algorithm is adopted to optimize the trainable parameters. The initial learning rate is set as 0.0001 and decreases with the increment of training epochs. Before training, data augmentation is employed by randomly dividing the images into 80x80 patches and rotating them by 0°, 90°, 180°, and 270°randomly. The batch size is set as 128.

Table 2. Color image denoising results of different networks

| Datasets | Methods | $\sigma = 15$ | $\sigma = 25$ | $\sigma = 50$ |
|---|---|---|---|---|
| Kodak24 | *FFDNet[36]* | 34.69 | 32.16 | 29.00 |
| | *DnCNN[35]* | 34.55 | 32.07 | 28.86 |
| | *ADNet[28]* | 34.76 | 32.26 | 29.10 |
| | *BRDNet[29]* | 34.89 | 32.44 | 29.22 |
| | *PAN-Net[23]* | **35.41** | **32.89** | 29.37 |
| | *MHCNN* | 35.12 | 32.63 | **29.46** |
| MCMaster | *FFDNet[36]* | 34.71 | 32.37 | 29.20 |
| | *DnCNN[35]* | 33.46 | 31.55 | 28.61 |
| | *ADNet[28]* | 33.99 | 31.31 | 28.04 |
| | *BRDNet[29]* | 35.10 | 32.77 | 29.52 |
| | *PAN-Net[23]* | **35.61** | **33.08** | 29.67 |
| | *MHCNN* | 35.35 | 33.01 | **29.83** |

Table 3. Results for different networks on real-world noise datasets.

| Test Data | **SIDD validation** | | | |
|---|---|---|---|---|
| Method | CBDNet[10] | RIDNet[2] | VDN[32] | MHCNN |
| PSNR | 38.68 | 38.71 | **39.28** | 39.06 |
| SSIM | 0.901 | **0.914** | 0.909 | **0.914** |
| Test Data | **DND** | | | |
| Method | CBDNet[10] | RIDNet[2] | VDN[32] | PAN-Net[23] | MHCNN |
| PSNR | 38.06 | 39.26 | 39.38 | 39.44 | **39.52** |
| SSIM | 0.942 | **0.953** | 0.952 | 0.952 | 0.951 |

## 4  Experimental Results

### 4.1  Datasets

MHCNN is tested on the tasks of AWGN denoising and real-world image denoising.

For AWGN denoising, the training set includes 400 images from [4], 400 images from the validation set of ImageNet [6] and 4,744 images from the Waterloo Exploration Database [22]. The AWGN noise generation algorithm from [35] is used to generate the noisy images. We train MHCNN on noise levels 15, 25, and 50, respectively, determined by the Gaussian distribution's standard deviation $\sigma$. For each noise level, MHCNN is tested on the commonly used datasets Set12 [26] for gray images and MCMaster [37] and Kodak24 [9] for color images.

We use the training set from the Smartphone Image Denoising DATA set (SIDD) [1] to train MHCNN for real-world image denoising. It includes 160 different scene instances, and each scene instance has two pairs of high-resolution images. Each pair includes one noisy image and its corresponding ground-truth image. In total, there are 320 training image pairs. The testing sets are the SIDD validation and the Darmstadt Noise Data set (DND) [25]. DND includes 50 pairs of images from four consumer cameras. There are no available ground-truth images online for DND, so we submit the denoising images to the DND official website to get the results.

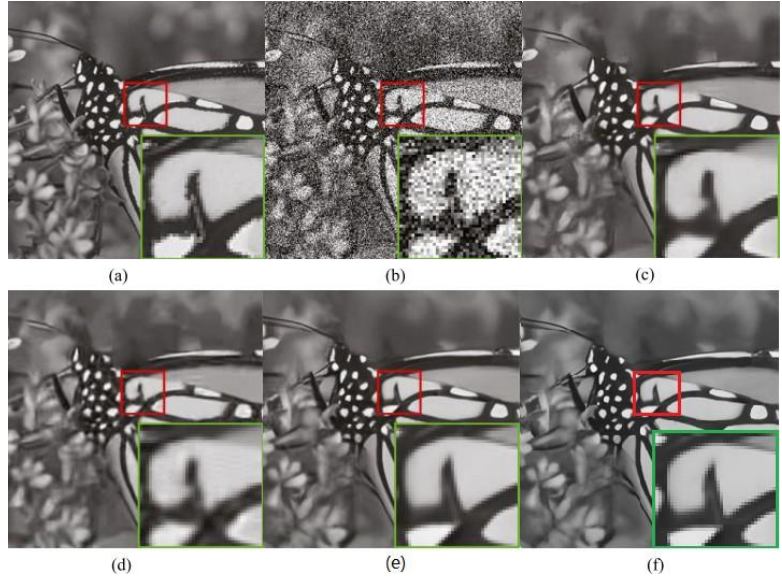

**Fig. 3.** Denoising results of the image Monarch from Set12 with noise level $\sigma$ = 50: (a) original image, (b) noisy image/14.71 dB, (c) DnCNN[35]/26.78 dB, (d) BM3D[5]/25.82 dB, (e) BRDNet[29]/26.97 dB, and (f) MHCNN/27.12 dB.

### 4.2  Comparison with other methods

**AWGN denosing** The peak signal-to-noise ratio (PSNR) [12] results are shown in Table 1 for gray images and Table 2 for color images. MHCNN performs better than other methods. At noise level $\sigma$ = 50, MHCNN gets the best PSNR results on every dataset, which shows it is powerful for high-level noise.

We visualize the denoising results of MHCNN and other models. Fig. 3 shows the denoising results of Monarch from Set12 at noise level $\sigma$ = 50. MHCNN removes the noise well. For color images, the denoising results of the apartment wall from Kodak24 are shown in Fig. 4. The wall contains rich details information, which is very suitable for evaluating the model's performance. The denoised image of MHCNN preserves the most details and has the best visual effects.

**Real-world image denoising** MHCNN has high performance on AWGN noise removing. We also do experiments on real-world images to ensure that MHCNN can be used in actual denoising tasks.

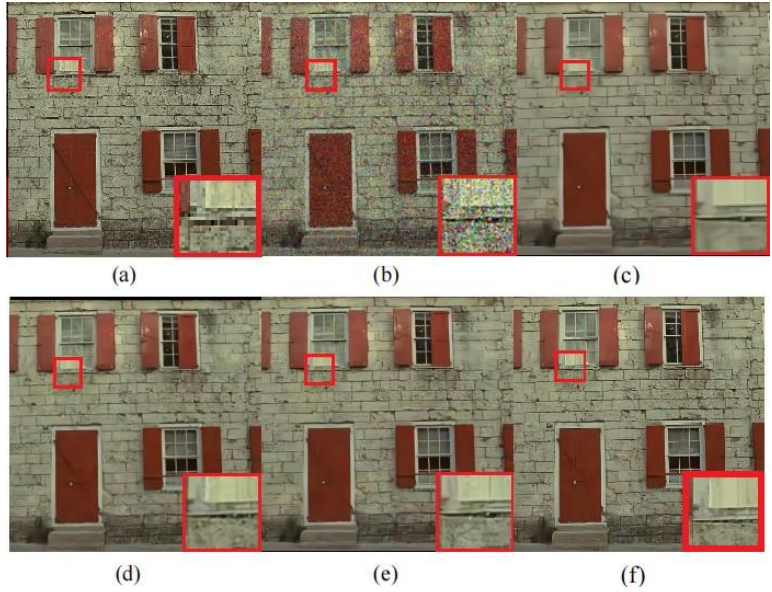

**Fig. 4.** Denoising results of the image from Kodak24 with noise level $\sigma = 50$: (a) original image, (b) noisy image, (c)DnCNN[35]/25.80dB, (d) BRDNet[29]/26.33dB, (e) FFDNet[36]/26.13dB, and (f) MHCNN/26.52dB.

Table 3 lists the results of different methods on SIDD validation and DND. MHCNN has the best Structure Similarity Index Measure (SSIM) [38] on SIDD and the best PSNR on DND, demonstrating its superiority. Fig. 5 shows the denoising result on DND, from which we observe that the noise has been removed successfully. Compared with other models, MHCNN does not smooth the local area of the noise too much and retains many details.

## 5  Results and Discussion

### 5.1  Ablation Experiments on MH

This section verifies the effectiveness of MH in MHCNN. Table 4 shows the PSNR results. We firstly study whether the number of heads affects denoising performance. According to Table 4, as the number of heads decreases, the denoising ability of MHCNN decreases. Of course, more heads will lead to more computation cost. We empirically choose three heads for a balance.

Different rotation angles of the input image are also studied. The results in Table 4 show that MHCNN (0°, 90°, 270°) and MHCNN (0°, 180°, 270°) have little impact on the performance. However, MHCNN (0°, 0°, 0°) leads to a performance degradation. This demonstrates that it is important to consider different rotation angles simultaneously.

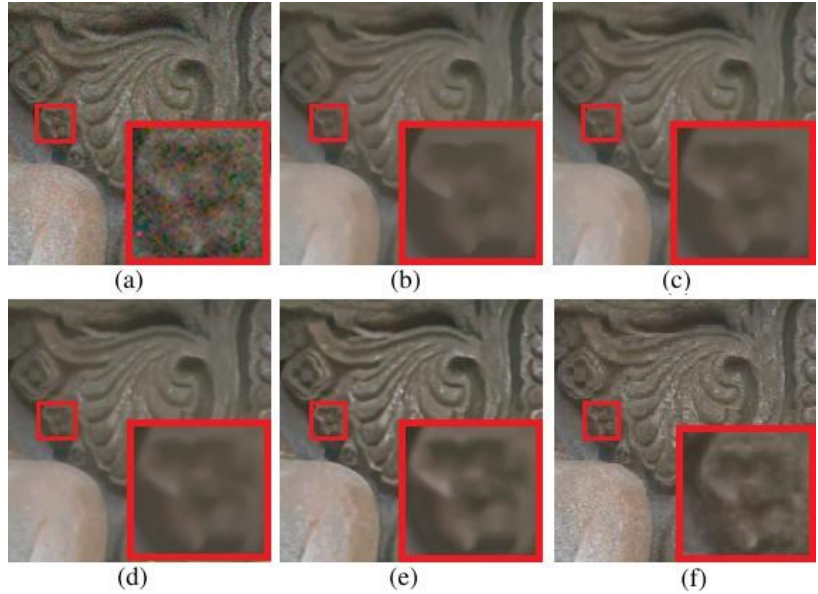

**Fig. 5.** Visual comparisons between MHCNN and other models. The test image was cropped from DND benchmark. (a) Input. (b) CBDNet[10]. (c) RIDNet[2]. (d) VDN[32]. (e) PAN-Net[23]. (f) MHCNN.

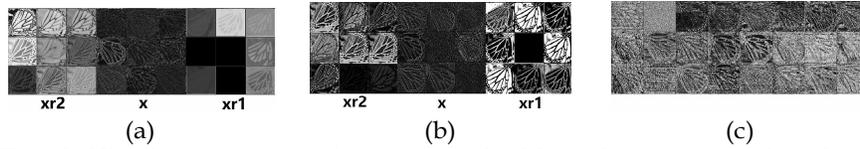

**Fig. 6.** Visualization of partial features of the Monarch. (a) is from three heads of MHCNN without MPA, (b) is from three heads MHCNN and (c) is the output of MPA.

### 5.2 Ablation Experiments on MPA

MPA achieves two crucial functions. One is integrating features from different heads of MHCNN. We visualized the features of the image Monarch extracted from each CNN head. As shown in Fig. 6 (b) and (c), the output features of MPA have the same angle as $x$, although the input angles are different. It illustrates that MPA projects the rotation images to the original image, achieving feature integration. The other is that MPA helps to extract rich features. Fig. 6 (a) and (b) show that MHCNN with MPA extracts richer images features, which have more details. For denoising performance, Table 4 shows MHCNN obtains higher PSNR results than that without MPA.

**Table 4.** Results (PSNR) of ablation experiments on Set12.

| Methods | $\sigma = 15$ | $\sigma = 25$ | $\sigma = 50$ |
|---|---|---|---|
| MHCNN | 33.21 | **30.84** | **27.70** |
| *MHCNN with 2 heads* | 33.15 | 30.75 | 27.63 |
| *MHCNN with 1 head* | 33.10 | 30.71 | 27.58 |
| *MHCNN (0°, 0°, 0°)* | 33.18 | 30.77 | 27.63 |
| *MHCNN (0°, 90°, 270°)* | **33.22** | 30.83 | **27.70** |
| *MHCNN (0°, 180°, 270°)* | 33.21 | 30.82 | 27.69 |
| *MHCNN without MPA* | 33.14 | 30.75 | 27.65 |

**Table 5.** Results (PSNR) of pluggable MH with MPA on Set12.

| Methods | $\sigma = 15$ | $\sigma = 25$ | $\sigma = 50$ |
|---|---|---|---|
| DnCNN[35] | 32.86 | 30.43 | 27.18 |
| *Multi-head DnCNN* | **32.97** | **30.56** | **27.32** |

### 5.3  Pluggable MH with MPA

We indicate that the MH with MPA mechanism is also valuable for the other single-head CNN models such as DnCNN. We use it to replace the first 10 layers of DnCNN to keep the network depth unchanged for a fair comparison. The PSNR results are shown in Table 5, which demonstrates that adding MH with MPA significantly improves DnCNN.

## 6  Conclusion

This paper proposes a novel denoising network named MHCNN. It has the start-of-the-art results for AWGN denoising and real-world image denoising. The MH with MPA mechanism in MHCNN is proved effective by the ablation experiments. In addition, this mechanism can also be added to other models to improve performance. Although MH will cause many parameters and calculations, excellent parallelism can solve this problem. MHCNN adopts three CNN heads to obtain image features to balance computing costs. Furthermore, more CNN heads can be used to improve the denoising effect in practical applications. MH with MPA is a valuable attention mechanism, and we will exploit MH with MPA in image recognition and other visual tasks in the future.

### Acknowledgment

This paper is supported by the National Natural Science Foundation of China (grant no. 62176241) and the National Key Research and Development Program of China (grant No. 2021ZD0200300) and the Open Project Program of the State

Key Laboratory of Mathematical Engineering and Advanced Computing(grant no. 2020A09).